%% file: main.tex
\documentclass[runningheads]{llncs}
\usepackage{graphicx}
\usepackage[numbers]{natbib}
\usepackage{amssymb}
\usepackage{enumitem}

\newcommand{\cardinality}[1]{\left\vert#1\right\vert}

\begin{document}
\title{Empirical Study of Easy and Hard Examples \\in CNN Training}
\author{Ikki Kishida \and
Hideki Nakayama}
\institute{The University of Tokyo, Graduate School of Information Science and Technology, Japan\\
\email{\{kishida,nakayama\}@nlab.ci.i.u-tokyo.ac.jp}}

\maketitle              % typeset the header of the contribution
\begin{abstract}
\input{abstract.tex}
\end{abstract}
\section{Introduction}
\input{introduction.tex}

\section{Method}
\input{method.tex}

\section{Experiments}

\subsection{Preparations}
\input{preparations.tex}

\subsection{Visual Property of Easy and Hard Examples}
\input{visualization_easy_hard.tex} 

\subsection{Are Easy and Hard Examples are common between different CNN architectures?}
\input{identical.tex}

\newpage
\subsection{Why some examples are consistently easy or hard?}
\input{why_easy.tex}

\subsection{Generalization and Easiness}
\input{ablation.tex}

\section{Related Work}
\input{related_work.tex}

\section{Conclusion}
\input{conclusion.tex}

\section{Acknowledgements}
\input{acknowledgement.tex}

\bibliographystyle{splncs04}
\bibliography{mybibliography}

\end{document}

%% file: abstract.tex
Deep Neural Networks (DNNs) generalize well despite their massive size and capability of memorizing all examples.
There is a hypothesis that DNNs start learning from simple patterns and the hypothesis is based on the existence of examples that are consistently well-classified at the early training stage (i.e., \textit{easy examples}) and examples misclassified (i.e., \textit{hard examples}).
Easy examples are the evidence that DNNs start learning from specific patterns and there is a consistent learning process.
It is important to know how DNNs learn patterns and obtain generalization ability, however, properties of easy and hard examples are not thoroughly investigated (e.g., contributions to generalization and visual appearances).
In this work, we study the similarities of easy and hard examples respectively for different Convolutional Neural Network (CNN) architectures, assessing how those examples contribute to generalization.
Our results show that easy examples are visually similar to each other and hard examples are visually diverse, and both examples are largely shared across different CNN architectures.
Moreover, while hard examples tend to contribute more to generalization than easy examples, removing a large number of easy examples leads to poor generalization.
By analyzing those results, we hypothesize that biases in a dataset and Stochastic Gradient Descent (SGD) are the reasons why CNNs have consistent easy and hard examples.
Furthermore, we show that large scale classification datasets can be efficiently compressed by using \textit{easiness} proposed in this work.
\keywords{Easy examples \and Hard examples \and Deep neural networks \and Dataset compression.}

%% file: introduction.tex
From a traditional perspective of generalization, overly expressive models can memorize all examples and result in poor generalization.
However, deep neural networks (DNNs) achieve an excellent generalization performance even if models are over-parameterized \citep{rethinking}.
The reason for this phenomenon remains unclear.
Arpit et al. \citep{closerlook} show that DNNs do not memorize examples, and propose a hypothesis that DNNs start learning from simple patterns.
Their hypothesis is based on the existence of examples that are consistently well-classified at the early training stage (i.e., easy examples) and examples misclassified (i.e., hard examples).
If DNNs memorize examples in brute force way, easy examples should not exist.
Easy examples are the evidence that DNNs start learning from specific patterns and there is a consistent learning process.
Therefore, we believe that analyzing easy and hard examples is one of the keys to understanding what kind of learning process DNNs have and how DNNs obtain generalization ability.

In this work, we study easy and hard examples, and their intriguing properties are shown.
For our experiments, we introduce \textit{easiness} as a metric to measure how early examples are classified correctly.
In addition, we calculate the matching rates of easy and hard examples between different CNN architectures.
As a result, we discover that both easy and hard examples are largely shared across CNNs, and easy examples are visually similar to each other and hard examples are visually diverse.

These results imply that CNNs start learning from a larger set of visually similar images and we hypothesize that easy and hard examples originate from biases in a dataset and Stochastic Gradient Descent (SGD).
A dataset naturally contains various biases leading some images to appear as a majority or a minority.
For instance, if there are many white dogs and rarely black dogs in dog images, the majority of visually similar images (i.e., white dogs) become easy examples and visually unique examples (i.e., black dogs) become hard examples.
Since SGD randomly picks samples for training a model, discriminative patterns in easy examples tend to be focused more than those in hard examples.
Thus, the gradient values of easy examples dominate the direction of the update at the beginning of training.
Such intra-class biases are the reason why some examples are classified well at an early training stage.

According to this hypothesis, the gradient values of easy examples are thought to be redundant and we may be able to remove easy examples without significantly affecting generalization ability.
To investigate how easy and hard examples contribute differently to generalization, we conduct ablation experiments.
We find that hard examples contribute more to generalization than easy examples, however, removing a large number of easy examples leads to poor generalization.
By using \textit{easiness}, we show that datasets can be efficiently compressed than random selection even in the large-scale ImageNet-1k dataset \citep{ILSVRC}.

\noindent Our contributions are as follows:
\vspace{-0.2cm}
\begin{itemize}[noitemsep,topsep=0pt]
    \item{We propose \textit{easiness} to measure how early an example is classified correctly}
    \item{Empirical finding and analysis of easy and hard examples based on \textit{easiness}. For instance, easy examples are visually similar to each other and hard examples are visually diverse, and both easy and hard examples are largely shared across different CNN architectures. We hypothesize such properties originate from the biases in the dataset and SGD.}
    \item{We demonstrate dataset compression by \textit{easiness}. It is more efficient than random selection and works even for the large-scale ImageNet-1k dataset.}
\end{itemize}

%% file: method.tex
\subsection{Easiness}
To measure how early an example is classified correctly, we introduce \textit{easiness} $e_{x_{i}}^T \in \mathbb{R}$ as a criterion, where $x_i$ represents one example and $T \in \mathbb{N}$ is the number of the model updates.
For a criterion of how correctly a model classifies the example, the loss value is appropriate.
However, since the model is stochastically updated, the loss value is uncertain in a single trial.
To improve the certainty of the loss value, it is necessary to take an average of the loss value over several times.
We propose \textit{easiness} $e_{x_{i}}^T$ that is the averaged loss value as follows:

\begin{equation}
    e_{x_i}^{T} = \frac{1}{M} \sum_{m=1}^M L(\textbf{t}_i, f(\textbf{x}_i, \textbf{W}_m^T)),
    \label{e}
\end{equation}
where $f(\textbf{x}_i, \textbf{W}_m^T)$ is the prediction and $t_i$ is the corresponding ground truth label.
$L$ is the loss function, for which we use the cross-entropy in this work since we focus on image classification.
$M \in \mathbb{N}$ is the number of trials and we set $M$ as $10$ in this work.

In this work, we define \textbf{10\% of the examples with the lowest \textit{easiness} as easy examples} and \textbf{10\% of the highest as hard examples}.

\subsection{Matching Rate}
It is important to know how large easy and hard examples are shared between various CNN architectures.
If easy and hard examples are not shared, it means that the learning process depends on the architecture of CNN and model-dependent analysis would be required.
To calculate the consistency of the set of examples, we use matching rate in this work.
Let us consider two different sets of examples $X_{A}$ and $X_{B}$. The matching rate $M_{AB} \in \left[0, 1\right]$ between $X_{A}$ and $X_{B}$ is calculated as

\begin{equation}
    M_{AB} = \frac{\cardinality{X_{A} \cap X_{B}}}{\max (\cardinality{X_{A}}, \cardinality{X_{B}})}, 
\end{equation}
where $\cardinality{\ \ }$ denotes the size of a set.

%% file: preparations.tex
We use CIFAR-10 \citep{cifar} and ImageNet 2012 dataset (ImageNet-1k) \citep{ILSVRC} for our experiments.

\noindent \textbf{CIFAR-10.}
CIFAR-10 is the image classification dataset.
There are 50000 training images and 10000 validation images with 10 classes.
For data augmentation and preprocessing, translation by 4 pixels, stochastic horizontal flipping, and global contrast normalization are applied onto images with $32 \times 32$ pixels.
We use three types of models of WRN 16-4 \citep{wideresnet}, DenseNet-BC 12-100 \citep{densenet} and ResNeXt 4-64d \citep{resnext}.

\noindent \textbf{ImageNet-1k.}
ImageNet-1k is the large scale dataset for the image classification.
There are 1.28M training images and 50k validation images with 1000 classes.
For data augmentation and preprocessing, resizing images with the scale and aspect ratio augmentation and stochastic horizontal flipping are applied onto images.
Then, global contrast normalization is applied to randomly cropped images with $224 \times 224$ pixels.
In this work, we use AlexNet \citep{alexnet}, ResNet-18 \citep{resnet}, ResNet-50 and DenseNet-121 \citep{densenet}.

As the optimizer, we use Momentum SGD with 0.9 momentum and weight decay of 0.0001.
The initial learning rate is 0.1 and it is divided by 10 at [150th, 250th] epochs and [100th, 150th, 190th] epochs on CIFAR-10 and ImageNet-1k, respectively.

%% file: visualization_easy_hard.tex
\begin{figure}[htpb]
\hspace{-0.9cm}
  \centering
    \begin{tabular}{c}
      \begin{minipage}{0.5\hsize}
         \centering
          \includegraphics[width=5cm]{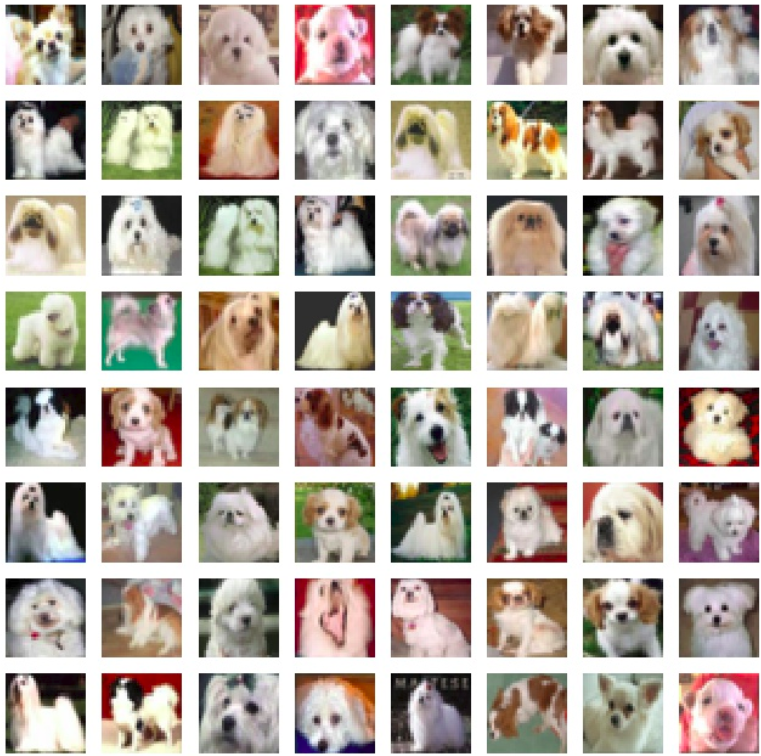}
          \hspace*{-0.0cm} \\(a) Easiest examples of dog
      \end{minipage}

      \begin{minipage}{0.5\hsize}
        \centering
          \includegraphics[width=5cm]{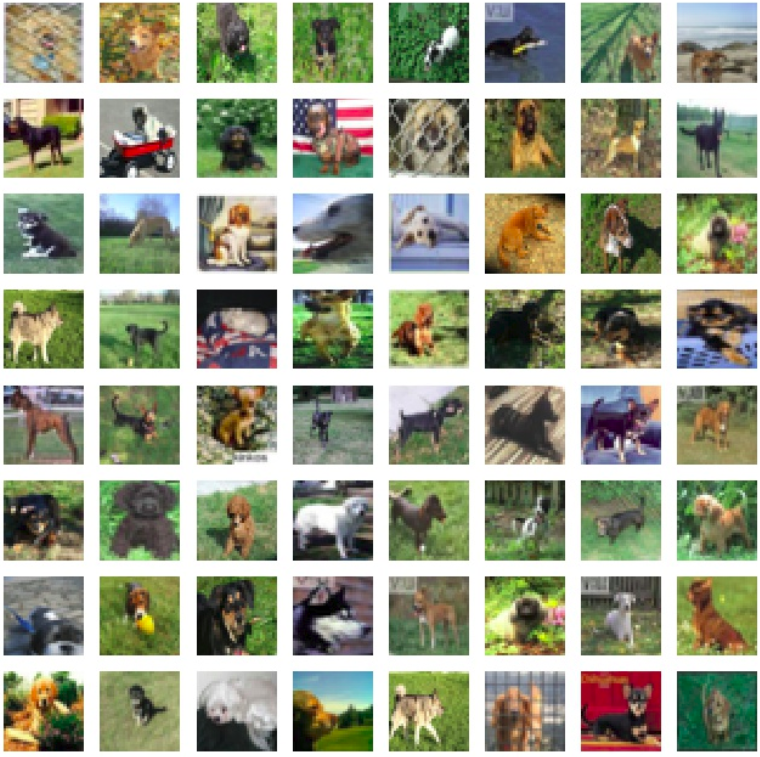}
          \vspace*{-0.0cm}\\(b) Hardest examples of dog
      \end{minipage}

    \end{tabular}
    \\
    \hspace{-0.9cm}
        \begin{tabular}{c}
      \begin{minipage}{0.5\hsize}
         \centering
          \includegraphics[width=5cm]{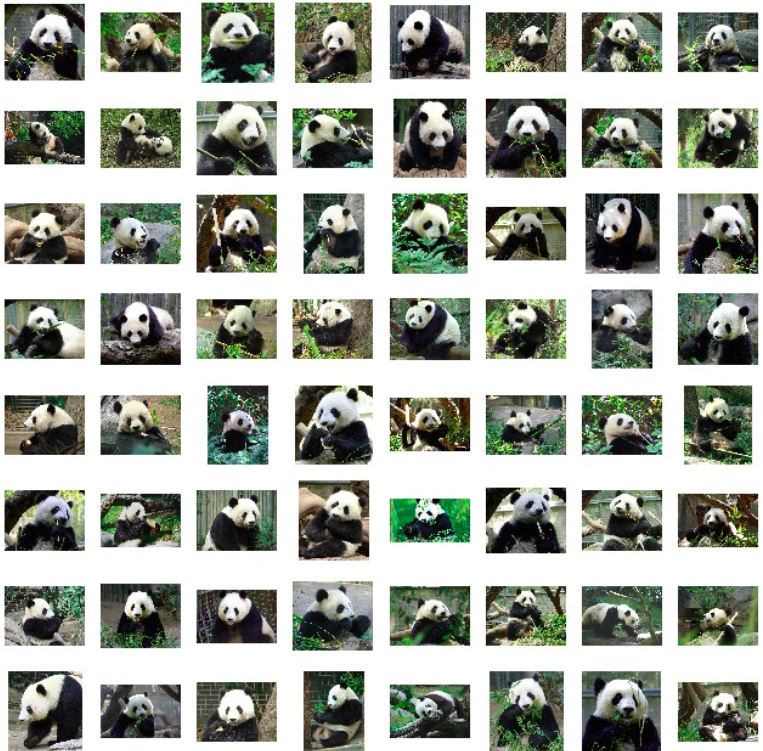}
          \hspace*{0.0cm}\\(c) Easiest examples of panda
      \end{minipage}

      \begin{minipage}{0.5\hsize}
        \centering
          \includegraphics[width=5cm]{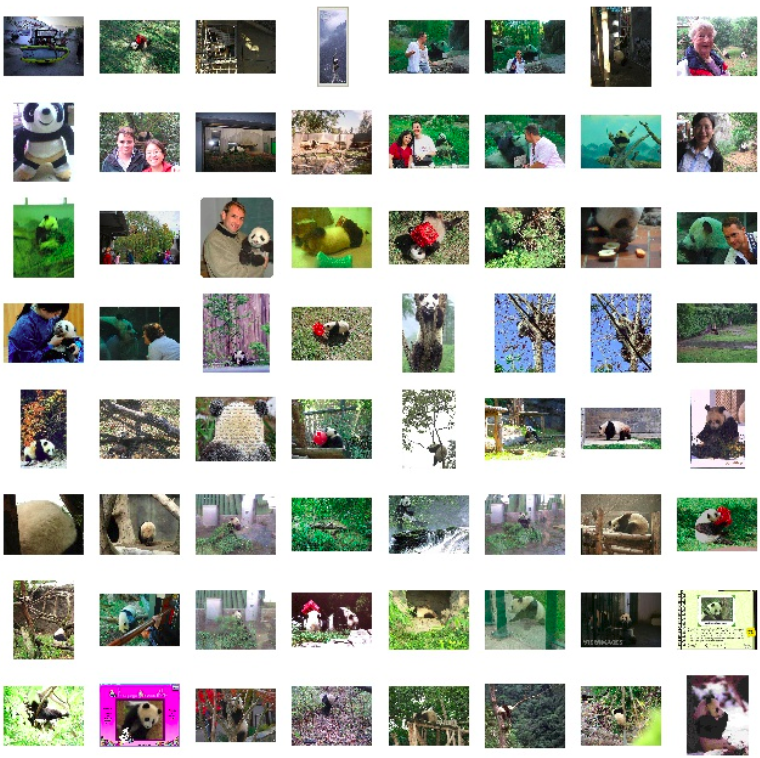}
          \hspace*{0.0cm}\\(d) Hardest examples of panda
      \end{minipage}

    \end{tabular}
    \caption{
        Easiest and hardest examples of CIFAR-10 and ImageNet-1k dataset.
        \textbf{a-b)} are from CIFAR-10 with \textit{easiness} of WRN 16-4.
        \textbf{c-d)} are from ImageNet-1k with \textit{easiness} of AlexNet.
     }
    \label{easy_hard_visualization1}
\end{figure}

Figure \ref{easy_hard_visualization1} shows easy and hard examples in CIFAR-10 and ImageNet-1k dataset.
Regardless of the size of the dataset, easy examples are visually similar to each other, and hard examples tend to be visually diverse.

\begin{figure}[htpb]
\hspace{-0.9cm}
  \centering
    \begin{tabular}{c}
      \begin{minipage}{0.33\hsize}
         \centering
          \includegraphics[width=4.0cm]{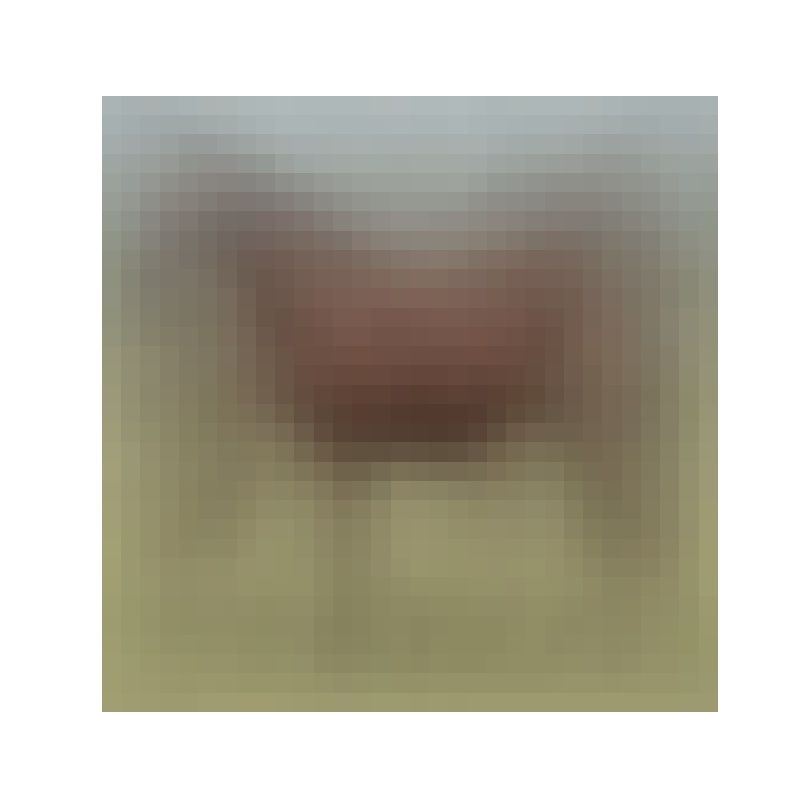}
          \hspace*{0.0cm}(a) \small{Averaged easy examples of horse}
      \end{minipage}

      \begin{minipage}{0.33\hsize}
        \centering
          \includegraphics[width=4.0cm]{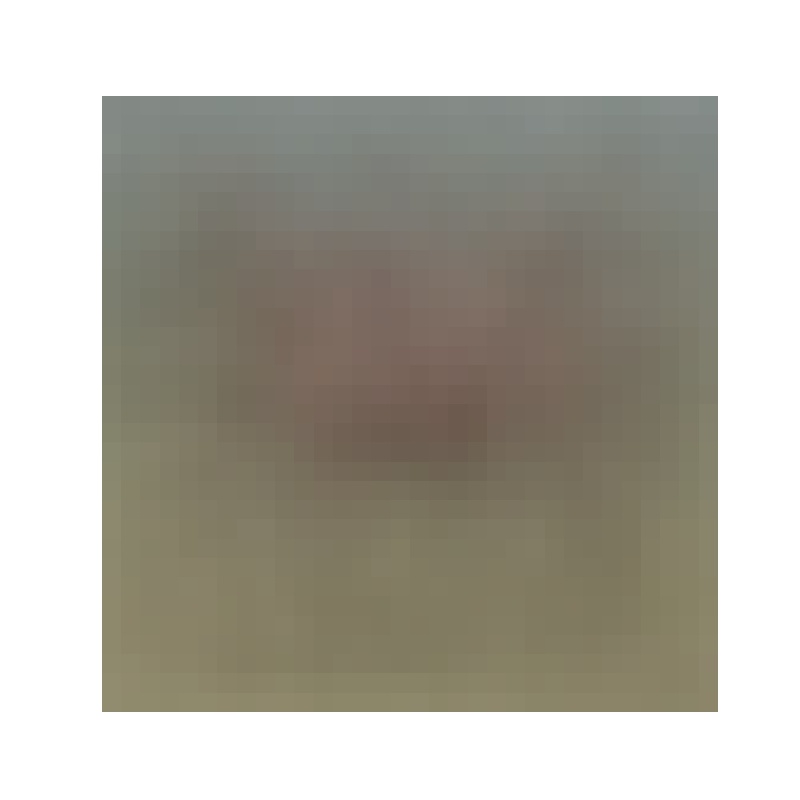}
          \hspace*{0.0cm}(b) \small{Averaged random examples of horse}
      \end{minipage}

      \begin{minipage}{0.33\hsize}
        \centering
          \includegraphics[width=4.0cm]{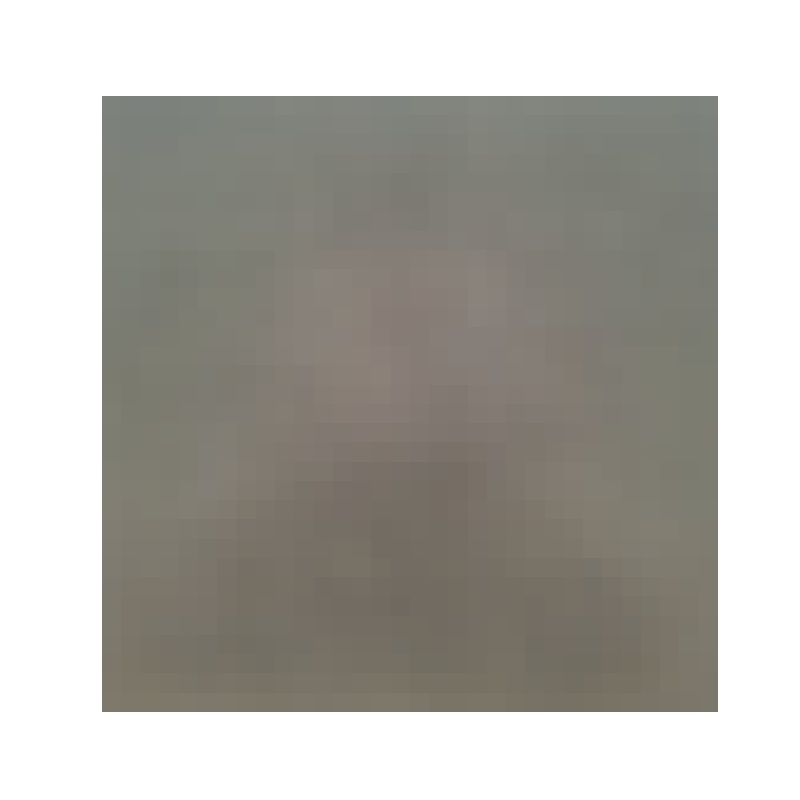}
          \hspace*{0.0cm}(c) \small{Averaged hard examples of horse}
      \end{minipage}
    \end{tabular}
    \caption{Averaged easy, random and hard examples of the horse in CIFAR-10. The \textit{easiness} is calculated by WRN 16-4. Each average image uses 500 images.}
    \label{easy_hard_visualization2}
\end{figure}

In \citep{datasetissue, imagenet_cvpr09}, the diversity of images is investigated by averaging the group of images.
The more diverse the images are, the more uniform the average image is.
The averaged images of easy and hard examples are shown in Figure \ref{easy_hard_visualization2}.
The averaged image of hard examples is more uniform than the averaged easy or random examples, thus hard examples are the most diverse among three.

Those results imply that CNNs start learning from a large set of visually similar images.

%% file: identical.tex
\begin{figure}[htpb]
\hspace{-0.9cm}
  \centering
    \begin{tabular}{c}
      \begin{minipage}{0.5\hsize}
         \centering
          \includegraphics[width=6.0cm]{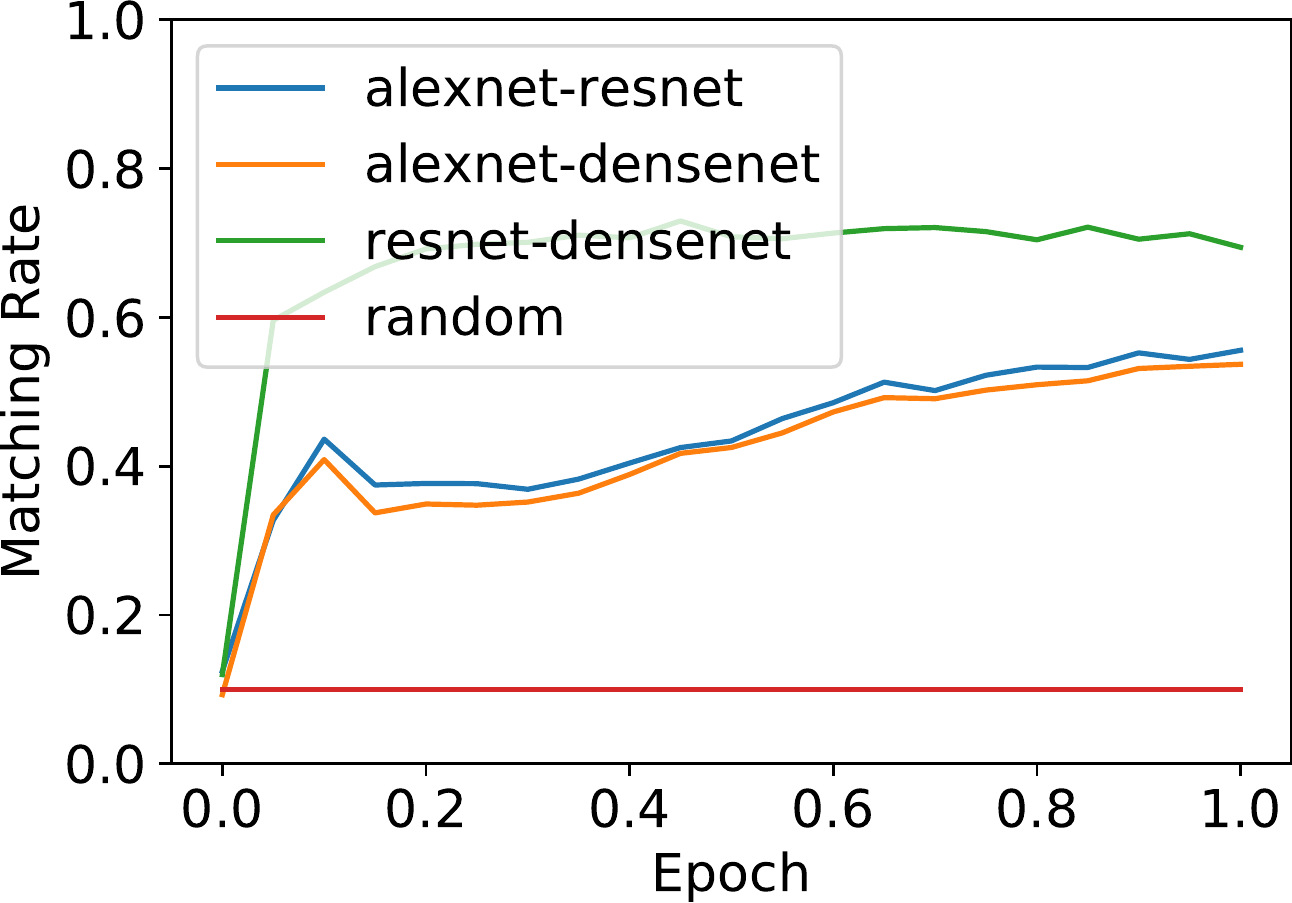}
          \hspace*{0.0cm}(a) \small{The matching rate of easy examples}
      \end{minipage}
      \begin{minipage}{0.5\hsize}
         \centering
          \includegraphics[width=6.0cm]{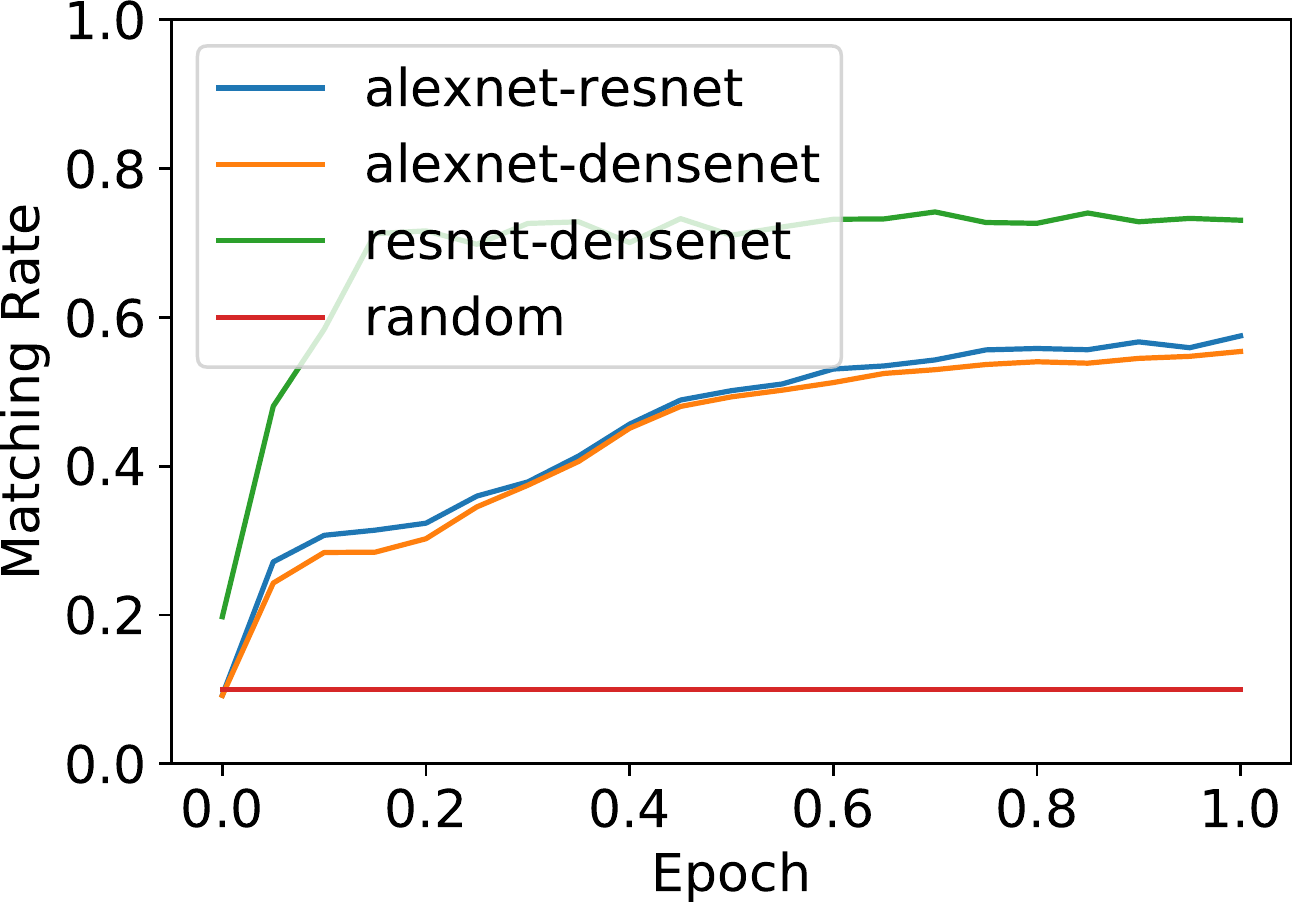}
          \hspace*{0.0cm}(b) \small{The matching rate of hard examples}
      \end{minipage}
    \end{tabular}
    \caption{
     The matching rate of easy and hard exmaples between different CNNs in ImageNet-1k. ``random" represents the chance rate of the case that 10\% of images are randomly sampled.}
    \label{easy_hard_visualization3}
\end{figure}

To investigate whether easy and hard examples are shared across different CNN architectures, we calculate matching rates according to \textit{easiness}.

Results are shown in Figure \ref{easy_hard_visualization3}.
The horizontal axis is the epoch and the vertical axis is the matching rate of easy and hard examples between different CNN architectures.
Easy and hard examples are largely shared at an early epoch and the matching rate is high across any architectures compared to random case. 
These results indicate that the learning process is similar regardless of the difference in the architecture design of CNN.

%% file: why_easy.tex
Results in previous experiments show that 
\begin{itemize}[noitemsep,topsep=0pt]
    \item{CNNs start learning from a larger set of visually similar images,}
    \item{Easy and hard examples are largely shared across different CNN architectures.}
\end{itemize}

We hypothesize that this phenomenon originates from dataset biases and Stochastic Gradient Descent (SGD). 

There are many biases in the dataset and \citep{unbiased} mentions several biases in a dataset.
\textit{Selection bias} means that examples in a dataset tend to have particular kinds of images (example: there are many examples of a sports car in the car category).
\textit{Capture bias} represents the manner in which photos are usually taken (example: a picture of a dog is usually taken from the front with the dog looking at the photographer and occupying most of the picture).
Easy examples are the result of such biases.

The parameters of CNN are updated by SGD based on calculated derivative values.
Since easy examples are visually similar to each other, it is expected that they get similar derivative values, and conversely, the derivative values of hard examples are unique. Therefore, the derivative values of easy examples are somewhat redundant. As a result, the derivative values of easy examples dominate the update of parameters at the beginning of learning.
From this learning process, we can explain why easy examples are classified well at an early stage, and easy and hard examples are common between different CNN architectures.

Arpit et al. \citep{closerlook} hypothesizes that CNN learns from simple patterns.
They measure the complexity of decision boundaries by Critical Sample Ratio (CSR).
CSR counts how many training examples are fooled by adversarial noises with radius $r$.
The higher CSR is, the more complex decision boundaries are.
Their results show that CSR becomes higher as CNN continues training, indicating that CNN firstly learns from simple patterns.
However, their hypothesis does not explain our results well such as the question of ``why the learned simple patterns are consistent between different CNNs".

Our hypothesis is the extension of Arpit et al. \citep{closerlook} in this respect.
We argue that CNNs firstly learn from simple patterns and such patterns are affected by the intra-class biases in a dataset.

%% file: ablation.tex
We perform ablation experiments on easy and hard examples to investigate if they equally contribute to the generalization ability.
For this purpose, we decide which examples to ablate based on \textit{easiness}.
In detail, we first normalize \textit{easiness} $e_{x_i}^T$ by dividing each $e_{x_i}^T$ by $\sum _{i=1}^N e_{x_i}^T$, where $N$ is the size of examples. We randomly select which to ablate by using the normalized \textit{easiness}.

\begin{figure}[htpb]
\hspace{-0.9cm}
  \centering
    \begin{tabular}{c}
      \begin{minipage}{0.5\hsize}
         \centering
          \includegraphics[width=6.0cm]{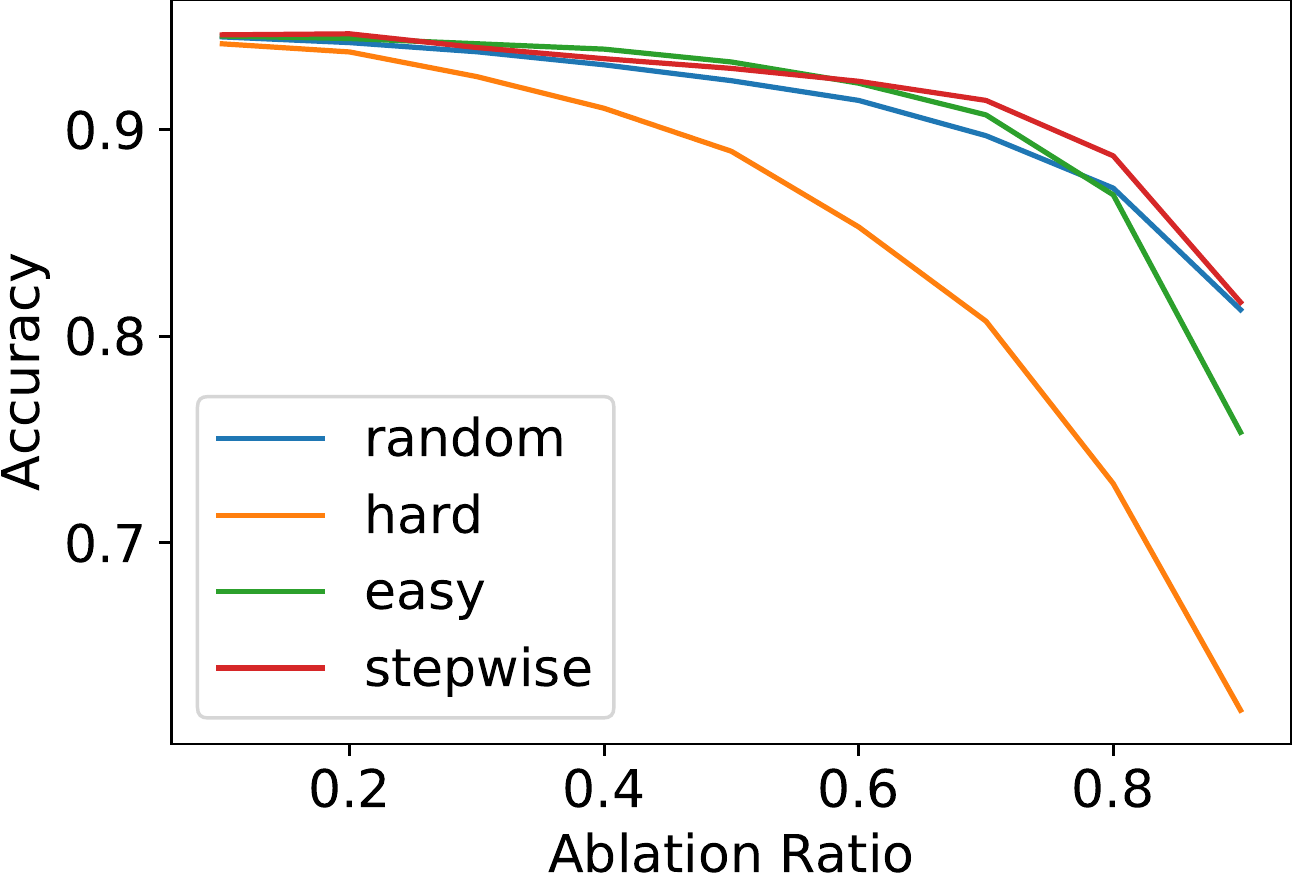}
          \hspace*{0.0cm}(a) \small{CIFAR-10}
      \end{minipage}
      \begin{minipage}{0.5\hsize}
         \centering
          \includegraphics[width=6.0cm]{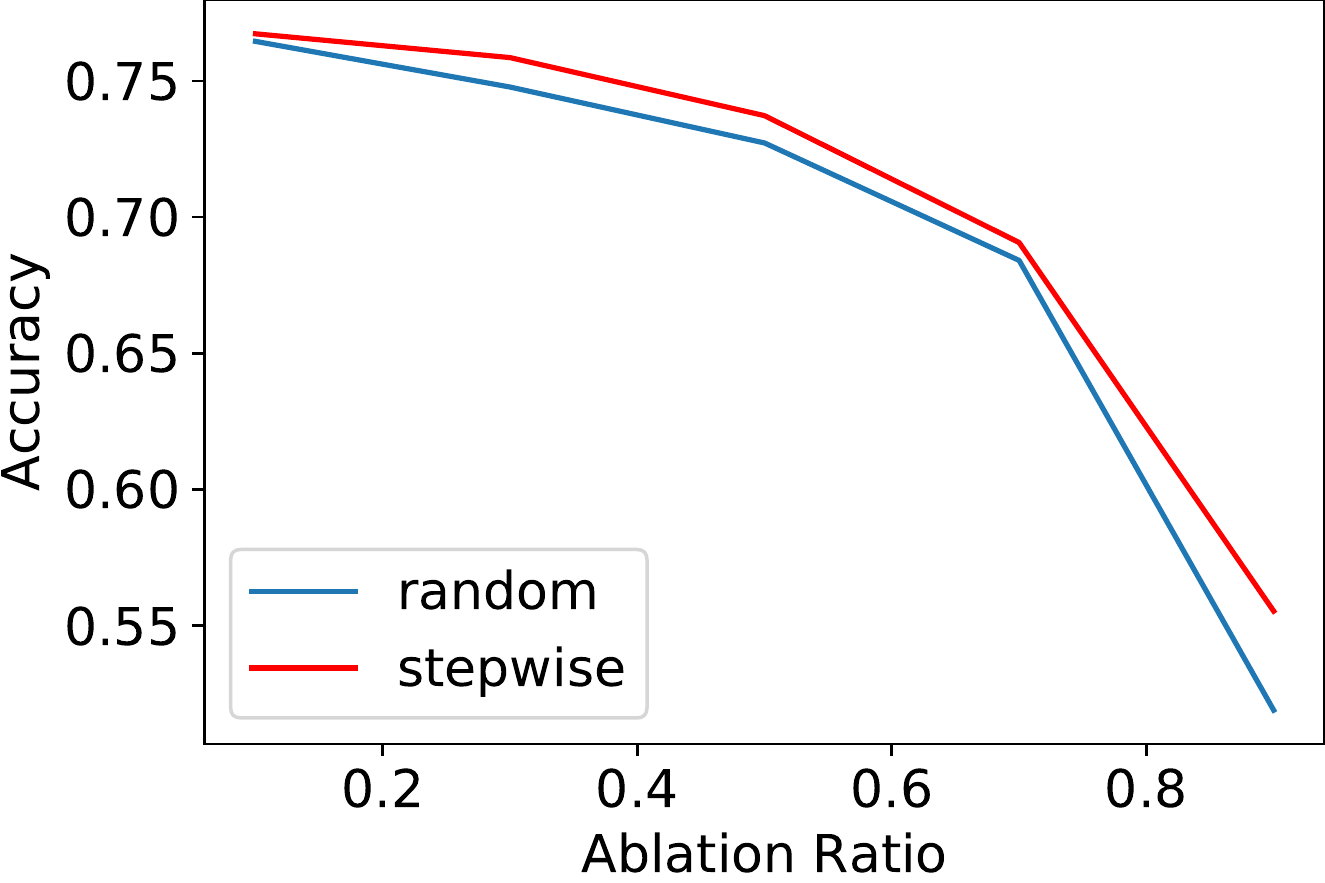}
          \hspace*{0.0cm}(b) \small{ImageNet-1k}
      \end{minipage}
    \end{tabular}
    \caption{
        The result of the ablation experiments.
        The vertical axis is accuracy and the horizontal axis is ablation ratio.
        If ablation ratio is 0.3, it means that 30\% of examples in the dataset are discarded.
        WRN 16-4 and ResNet-50 are used respectively for CIFAR-10 and ImageNet-1k dataset.
        ``easy", ``hard" and ``random" on figures means easy, hard and randomly selected examples are mainly removed, respectively.
        ``stepwise" is the gradual case of ``easy".
        Unlike "easy" that ablating examples in one shot, every time ``stepwise" ablates 10\% of easy examples in the dataset and re-calculate \textit{easiness} until reaching the target ablation ratio.
    }
    \label{result_random_ablation}
\end{figure}

The result is shown in Figure \ref{result_random_ablation}-(a,b).
The horizontal axis is the ablation ratio and the vertical axis is accuracy.
If ablation ratio is 0.3, then the size of the training dataset is 70\%.
 ``easy", ``hard" and ``random" on figures means easy, hard and randomly selected examples are mainly removed, respectively.
``stepwise" is the gradual case of ``easy".

As can be seen in Figure \ref{result_random_ablation}-a, removing hard examples consistently degrades the classification performance more drastically than other strategies.
Therefore, we conclude that hard examples contribute more to generalization than easy examples do.

However, as can be seen in ``easy" of Figure \ref{result_random_ablation}-a, if we remove too many easy examples, the accuracy starts degrading sharply.
This phenomenon can be explained by our hypothesis.
Since a dataset is randomly split into training and testing subsets, training and testing share the same biases.
For example, if white dogs are easy examples and black dogs are hard examples in the training dataset, there are more white dogs than black dogs in the testing dataset too.
Thus, if the trained model fails to learn white dogs (i.e., easy examples), the test accuracy will drop sharply since there are many white dogs in the testing dataset.
Therefore, it is better to keep some easy examples even though redundant images can be ablated with less affecting generalization ability.

``stepwise" keeps some of the easy examples while ablating them.
As can be seen in Figure \ref{result_random_ablation}, ``stepwise" gives the best performance.
In addition, in Figure \ref{result_random_ablation}-b, ``stepwise" outperforms ``random" case even in the large-scale ImageNet-1k dataset.
The difference in accuracy between ``random" and ``stepwise" is around 1.1\% at 0.3 ablation ratio.
It is approximately worth extra 100k images to achieve the comparable accuracy in ``random".

%% file: related_work.tex
A dataset naturally contains various biases.
For instance, Ponce et al. \citep{datasetissue} shows some averaged images of Caltech-101 \citep{caltech101} are not homogeneous and recognizable.
They claim that Caltech-101 may have \textit{inter-class} variability but lacks \textit{intra-class} variability.
In this work, we find that easy examples lack \textit{intra-class} variability, and hard examples are more diverse than easy examples.

%There are three types of training schemes that start learning from easy examples \citep{curriculum, selfpace}, uncertain examples\citep{activelearningsurvey,activebias} or hard examples \citep{focalloss,fracking,ohem,batchselection,unsupervisedvideo}.
%Those methods mainly rely on the loss value for each training example. Thus properties of easy and hard examples are implicitly used.
%For instance, when hard examples are emphasized, hard examples are learned at an early stage. As a result, imbalanced problems in \textit{intra-class} and \textit{inter-class} are mitigated.

Arpit et al. \citep{closerlook} investigate the memorization of DNNs, and claims that DNNs tend to prioritize learning simple patterns first.
They analyze the complexity of the decision boundary based on Critical Sample Ratio (CSR).
CSR is the criterion of how many training examples change the predictions by adding adversarial noises with radius $r$.
The high CSR means that CNN has complex decision boundaries.
Arpit et al. \citep{closerlook} empirically show that CSR becomes higher as CNNs continue training and propose the hypothesis that CNN learns from simple patterns.
However, their hypothesis does not explain why firstly learned simple patterns are consistent between different CNN architectures.
Our hypothesis is the extension of \citep{closerlook} in this respect.
We argue that CNN firstly learns from simple patterns and such patterns are affected by the intra-class biases in a dataset.

Agata et al. \citep{valuable} investigate well-classified examples and misclassified examples at the end of training based on SVMs with hand-crafted features in the small-scale datasets.
They conclude that some examples are the reason to degrade the model's performances and examples with high loss values contribute to generalization well.
In this work, we empirically investigate and analyze properties of examples at an early training stage in CNNs, and perform experiments on the large-scale ImageNet-1k dataset.

Toneva et al. \citep{forgetting} empirically investigate forgettable and unforgettable examples on small-scale MNIST \citep{mnist} and CIFAR-10 dataset.
The difference of the metric between \citep{forgetting} and our \textit{easiness} is to use the loss values of the model with small updates unlike tracking the degradation of the accuracy across the whole training period in \citep{forgetting}.
Table \ref{table_matching_rate} shows the matching rates between easy and hard examples at the beginning and at the end of the training.
As can be seen in Table \ref{table_matching_rate}, easy and hard examples are different, especially in ResNeXt.
Therefore, we assume easy and hard examples are different from forgettable and unforgettable examples since \citep{forgetting} use records of the whole training period.

\begin{table}[t]
\centering
  \begin{tabular}{cccc}
  \hline\noalign{\smallskip}
  \small{model} & \small{easy or hard} & \small{matching rate}\\
  \hline  \noalign{\smallskip}
  WRN 16-40& Easy & 0.18\\
   & Hard & 0.255\\ \cline{2-2} \noalign{\smallskip}
  DenseNet-BC 12-100& Easy & 0.24\\
   & Hard & 0.279\\ \cline{2-2} \noalign{\smallskip}
  ResNeXt 4-64d& Easy & 0.129\\
   & Hard & 0.174\\
  \hline
  \end{tabular}
\caption{The matching rates between easy and hard examples at the beginning and at the end of training in CIFAR-10. The chance rate is $0.1$.}
\label{table_matching_rate}
%%%%%%%%%%%%%%%%%%%%%%%%%%%%
\end{table}

%% file: conclusion.tex
In this work, easy and hard examples are investigated to understand the learning process of DNNs.

Firstly, the metric of \textit{easiness} is introduced to define easy and hard examples. Then, we discover that easy and hard examples are common among different CNN architectures, and easy examples are visually similar to each other and hard examples are visually diverse.
To explain these phenomena, we propose the hypothesis that biases in the dataset and SGD make some examples easy or hard.

From this hypothesis, we consider easy examples are visually redundant and can be removed without significantly affecting the generalization ability of a model.
In ablation experiments, we demonstrate that hard examples contribute more to generalization ability than easy examples in CIFAR-10 and the large-scale ImageNet-1k dataset.
Therefore, the dataset can be efficiently compressed than random selections by using \textit{easiness}.

For future work, further analysis of intra-class biases is fruitful directions. In addition, studying how to design biases in a dataset is promising directions to control the learning process of CNNs.

%% file: acknowledgement.tex
This work was supported by JSPS KAKENHI Grant Number JP19H04166.
We would like to thank the two anonymous reviewers for their valuable feedback on this work.